%% file: main.tex
\documentclass[conference]{IEEEtran}

\usepackage{cite}
\usepackage{amsmath,amssymb,amsfonts}
\usepackage{graphicx}
\usepackage{textcomp}
\usepackage{xcolor}
\usepackage{booktabs}
\usepackage{verbatim}
\graphicspath{{figures/}}
\usepackage[singlelinecheck=false, justification=raggedright]{caption} 

\makeatletter
\g@addto@macro\normalsize{%
  \setlength\abovedisplayskip{4pt plus 1pt minus 1pt}%
  \setlength\belowdisplayskip{4pt plus 1pt minus 1pt}%
  \setlength\abovedisplayshortskip{2pt plus 1pt}%
  \setlength\belowdisplayshortskip{2pt plus 1pt}}
\makeatother
\begin{document}

\title{Fixed-Budget Gaussian Volume Encoding with Structure-Aware Allocation}

\author{\IEEEauthorblockN{Michael R. Martin}
\IEEEauthorblockA{\textit{Department of Computer Science} \\
\textit{University of California, Davis}\\
csemartin@ucdavis.edu}
\and
\IEEEauthorblockN{Joseph Insley\\
Victor A. Mateevitsi\\
Silvio Rizzi
}
\IEEEauthorblockA{\textit{Argonne National Laboratory}\\
insley, vmateevitsi, srizzi@anl.gov
}
\and
\IEEEauthorblockN{Kwan-Liu Ma}
\IEEEauthorblockA{\textit{Department of Computer Science} \\
\textit{University of California, Davis}\\
klma@ucdavis.edu}
}

\maketitle

\begin{abstract}
Scientific simulations often produce scalar volumes faster than they can be stored, transferred, and loaded, while in situ reduction must use only a limited share of simulation resources. This work encodes scalar fields as anisotropic Gaussian primitives under a fixed budget. The complete primitive set is allocated analytically from local field structure, including position, orientation, and shape, then refined directly against the scalar field without densification, pruning, or count changes. The selected budget determines encoded storage before refinement and, together with the iteration schedule, provides a controllable refinement-time budget. In a controlled benchmark, truncation-aware field evaluation reduces encoding time by up to $51\times$; 1.4 million Gaussians encode a billion-voxel volume in at most four minutes on one desktop GPU, with reduced-iteration refinement completing in under one minute. Across five datasets spanning 2.1 million to 1.1 billion evaluated voxels, compression-useful configurations achieve 15.0--38.7\,dB PSNR at compression ratios from $2.2\times$ to over $40{,}000\times$. Pre-encoding structure statistics characterize fields for which one-shot allocation yields limited gains from additional capacity. Because primitives retain scalar attributes rather than baked appearance, a single compact model serves every subsequent visualization state---supporting post-hoc transfer-function, colormap, lighting, and viewpoint changes without re-encoding.
\end{abstract}




\input{sections/01_introduction.tex}
\input{sections/02_related_work.tex}
\input{sections/03_method.tex}
\input{sections/04_experimental_setup.tex}
\input{sections/05_results.tex}
\input{sections/06_discussion.tex}
\input{sections/07_conclusion.tex}

\bibliographystyle{IEEEtran}
\bibliography{references}

\end{document}

%% file: sections/01_introduction.tex
\section{Introduction}

Scientific visualization relies on volumetric inspection of scalar fields whose structures of interest are spatially distributed, multiscale, and difficult to characterize from slices or isolated viewpoints~\cite{sarton2023sota,liu2025coordinated}; across domains such as fluid simulation, combustion, medical imaging, and computational science~\cite{fischer2021nekrs}, these structures are revealed through transfer functions, and direct volume rendering remains the standard reference because it preserves semitransparent internal structure~\cite{kniss2002multidimensional,max1995optical}. Dense grids, however, grow with simulation resolution---a single $1024^3$ float32 timestep occupies 4.3\,GB---and rendering costs repeated sampling, interpolation, and compositing after each transfer-function or camera update~\cite{morrical2023quick,bauer2023fovolnet}. On HPC systems the bottleneck starts earlier: full-resolution fields must be written, transferred, and loaded before analysis can begin, motivating effective data reduction for improved storage and computation management---whether applied in situ during simulation or post hoc after raw files are saved to disk~\cite{ma2009insitu,lindstrom2014zfp,antic2026}. Because the computing resources of a run are reserved for the simulation itself, an in situ reduction stage must operate within a small share of the run's cost---a common rule of thumb allows it no more than 5\%~\cite{ma2009insitu,yu2010insitu}. A fixed-bound stated in advance can only be enforced by an encoder that knows its cost in advance, which is what a specified budget provides and a growth-driven one cannot accomodate. Learned scene representations offer an alternative approach: neural fields and graphics primitives encode scenes through image-space supervision~\cite{mildenhall2020nerf,yu2021plenoxels,muller2022instant}, which have been applied via learned encodings to volumetric scalar data ~\cite{lu2021compressive,weiss2022fast,wu2024interactive}, where 3D Gaussian Splatting shows that explicit anisotropic primitives support real-time rendering while remaining spatially localized~\cite{kerbl2023gaussian}. This raises the question considered here: whether scientific scalar volumes can be represented by trainable Gaussian primitives that preserve scalar distributions and feature structure, carried by \emph{analytic, structure-aware allocation} rather than iterative densification. Unlike standard RGB radiance fields, scientific scalar fields require the retention of scalar attributes within the gaussian primitives for dynamic transfer-function evaluation~\cite{dyken2025veg} while adhering to imposed pre-allocated memory budgets~\cite{lu2021compressive}. Our evaluation addresses these constraints and focuses on: \textbf{(1)} a scalar-aware anisotropic Gaussian representation that enforces deterministic memory bounds by analytically fixing primitive cardinality prior to refinement; \textbf{(2)} analytic structure-aware allocation that commits the entire capacity from field statistics maintained prior to training, with a truncation-aware evaluation scheme cutting training cost up to $51\times$ and permitting million-primitive budgets supported outside traditional HPC environments; \textbf{(3)} a five-dataset evaluation of quality, compression, cost, and memory across $10^3$ to $1.4\times10^6$ primitives, with inexpensive pre-encoding structure metrics that characterize allocation demands ahead of training. 

%% file: sections/02_related_work.tex
\section{Related Work}

Direct volume rendering, in which a transfer function maps values to optical properties composited along rays~\cite{levoy1988display}, remains the reference for inspecting scalar volumes; hardware acceleration and specialized traversal reduce its rendering cost~\cite{kruger2003acceleration,laur1991hierarchical}, leaving the storage and movement of the dense grid as the problem an encoded representation addresses.

\subsection{In Situ Data Reduction and Visualization}
\label{sec:insitu-rw}

Extreme-scale simulations can produce more data than can be practically stored or transferred for post hoc analysis. In situ processing operates on data still resident in simulation memory under strict cost discipline---visualization and reduction are expected to take only a small fraction of the run, whose resources belong to the simulation~\cite{ma2009insitu,yu2010insitu}. Effective reduction matters beyond the in situ setting as well: when scientists dump data directly, a compact encoding still determines how much can be stored and how efficiently it is managed afterward. Prior reduction spans error-bounded lossy compressors~\cite{lindstrom2014zfp,di2016sz}, in situ video encoding on dedicated hardware~\cite{leaf2017video}, trigger-driven declarative workflows~\cite{wu2020diva}, distributed neural representations trained beside the simulation~\cite{wu2022distributed,wu2024distributed}, temporally adaptive in situ neural compressors~\cite{antic2026}, and Gaussian-splatting pipelines for high-resolution scientific visualization~\cite{han2025distributed}.

\subsection{Neural Scene and Volume Representations}

Neural representations parameterize a continuous signal using trainable networks or compact feature structures~\cite{sitzmann2020siren}: NeRF and its variants optimize density and radiance from posed images~\cite{mildenhall2020nerf,yu2021plenoxels}, and multiresolution hash encodings cut their optimization and rendering cost~\cite{muller2022instant,wu2024interactive}. Applied to volume data~\cite{wang2023dl4scivis}, learned encodings deliver compact storage and interactive rendering~\cite{weiss2022fast,lu2021compressive,bauer2023fovolnet,wu2024distributed}. However, unlike explicit primitive representations, they generally do not expose individually addressable renderable elements for primitive-level ranking or processing, and their storage footprint is determined by the selected network or encoding architecture rather than by a direct count of renderable elements. This distinction motivates explicit representations when an application requires a prescribed, readily interpretable storage budget or primitive-level control.

\subsection{Gaussian Splatting and Explicit Primitive Representations}

3D Gaussian Splatting represents a scene using anisotropic Gaussian primitives optimized through differentiable rendering, with density control that splits and prunes primitives during optimization~\cite{kerbl2023gaussian}. Subsequent work has extended the representation to large-scale reconstruction and textured appearance~\cite{xu2024grm,chao2025textured}, and has investigated compressed and hybrid neural--explicit variants for reducing model storage~\cite{tang2026econgs}. In scientific visualization, Gaussian-based methods have also been developed for editable volume exploration and accelerated volume rendering~\cite{tang2025ivrgs,bauer2026gscache}. A direct application to scientific data is nonetheless limited, as the optimized quantities are appearance parameters rather than scalar attributes to which a transfer function can be applied. Volume Encoding Gaussians (VEG) adapt Gaussian primitives to scientific volume rendering: Gaussians encode scalar values, color and opacity are assigned at render time through a transfer function, and the primitives are optimized through a differentiable splatting path against rendered images~\cite{dyken2025veg}. A subsequent formulation, W-VEG, shows that image-space-trained models do not transfer reliably to direct sampling, and moves supervision into world space (querying the field directly at 3D positions through normalized Gaussian kernel regression), where its primitive set is grown through error-guided densification until a target compression ratio is reached~\cite{dyken2026wveg}. The transfer function and lighting (a per-primitive Blinn--Phong term~\cite{blinn1977models}) are applied once per primitive and $\alpha$-blended in depth order, leading to point-sampled classification at primitive centers rather than along rays---motivating the separation of encoded field from rendering path (Section~\ref{sec:method}).

\subsection{Gaussian Count and Capacity Allocation}

In densification-based pipelines, density control is part of the optimization: primitives are split, inserted, or pruned as training proceeds~\cite{kerbl2023gaussian,dyken2026wveg}, so the final count emerges from the growth process and its steering machinery is paid for at every iteration. Uncapped model growth follows the error signal over predefined specifications, involving a standalone focus area into 3D Gaussian minimization~\cite{fan2024lightgaussian,niedermayr2024compressed,papantonakis2024reducing}. A target compression ratio can bound the final model size, but the bound is reached rather than set: insertion and pruning continue until the target ratio is met and loss converges, so training length depends on the data and the intended size is realized only once training ends~\cite{dyken2026wveg}.

\subsection{Positioning of This Work}

Across this prior work, capacity is either learned implicitly through architecture choice~\cite{weiss2022fast,muller2022instant} or grown in response to reconstruction error~\cite{tang2026econgs,kerbl2023gaussian,dyken2026wveg}; both make encoded size a consequence of optimization and offer no way to determine beforehand whether a field will be well served. A third position---deciding allocation analytically in advance, from the structure of the field itself, and holding it fixed---is to our knowledge unexplored for scientific scalar volumes, and is the position an in situ allowance requires. The remainder of this paper develops and evaluates it.

%% file: sections/03_method.tex
\section{Methodology}
\label{sec:method}

\subsection{Overview}

Prior scalar-aware Gaussian encoders adapt spatial capacity through growth and pruning toward a target compression ratio~\cite{dyken2025veg,dyken2026wveg}. Our study instead addresses this by treating primitive count as a fixed budget prior to training. The pipeline has three stages: an analytic structure analysis allocates the full budget in one pass before error signal consideration---positions drawn from a mixture of gradient-weighted and uniform distributions over the occupied domain, orientation and anisotropic shape from a local structure tensor, allowing allocation adequacy to be measured upfront (Section V-F); camera-free training then adjusts every parameter against field-space objectives at stratified sample points, with densification excluded and the count unchanged; the full-precision representation is exported at 48 bytes per primitive, with an optional 8-bit variant; the active transfer function assigns color and opacity at render time.

\subsection{Volume Preprocessing}

The evaluated implementation accepts a structured scalar volume on a regular voxel grid; the Gaussian representation itself uses unconstrained primitive positions. Input values are converted to float32 and linearly normalized to $[0,1]$ using the field minimum and maximum~\cite{dyken2026wveg}. Seeding and refinement operate in normalized-value and voxel-index coordinates, so all reported reconstruction errors refer to the normalized field. Compression ratios are measured against the source byte count at the evaluated resolution (Table~\ref{tab:datasets}), using the source's native storage format rather than its float32 in-memory representation.

\subsection{Scalar-Aware Gaussian Primitives}

Let the encoded volume be represented by a set of $N$ Gaussian primitives,
\begin{equation}
\mathcal{G} = \{g_i\}_{i=1}^{N}.
\end{equation}
Each primitive is parameterized as
\begin{equation}
g_i = (\mu_i, s_i, q_i, a_i, w_i),
\end{equation}
where $\mu_i \in \mathbb{R}^3$ is its center, $s_i \in \mathbb{R}_{>0}^3$ its anisotropic scales, $q_i$ a unit-quaternion rotation, $a_i$ its scalar attribute, and $w_i$ its weight. The scalar attribute is retained as a data attribute rather than converted directly into RGB appearance. At render time, the active transfer function maps the reconstructed scalar field to color and opacity, while $w_i$ modulates the primitive's contribution to the field. Each primitive is optimized in the standard unconstrained parameterization of 3DGS-family encoders (log-scales, normalized quaternion, sigmoid-activated logits)~\cite{kerbl2023gaussian,dyken2026wveg}. The Gaussian set therefore represents a scalar field whose appearance is assigned at render time. We reconstruct this field using density-normalized Gaussian kernel regression:
\begin{equation}
\tilde{V}(p) \;=\; \frac{\sum_i w_i\, a_i\, G_i(p)}{\max\!\left(\sum_i w_i\, G_i(p),\; \rho_0\right)},
\label{eq:field}
\end{equation}
where $G_i(p) = \exp\!\big(-\tfrac{1}{2}\, d_i(p)^\top \Sigma_i^{-1} d_i(p)\big)$ with $d_i(p) = p - \mu_i$ and $\Sigma_i$ the covariance induced by $(s_i, q_i)$. The kernel is truncated at $5\sigma$ (squared Mahalanobis distance below 25), and the density floor $\rho_0 = 0.05$ prevents unstable normalization in sparsely covered regions, where $\tilde{V}(p)$ approaches zero.

\subsection{Direct 3D Supervision}

Both allocation and training operate directly in voxel coordinates against sampled scalar values, as in world-space scalar Gaussian encoding~\cite{dyken2026wveg}: no cameras, rendered images, or transfer functions participate, and representation is independent from cross-dataset training---each volume optimizes a self-contained primitive set initialized from its own field structure. Rendering applies a transfer function $T=(c_T,\alpha_T)$ to the value the reconstructed field of Eq.~\ref{eq:field} returns, compositing those values along each ray. The encoded parameters $\{g_i\}$ therefore enter only through $\tilde{V}$, and $T$ acts as an outer mapping on its output; since the objective of Section~\ref{sec:objective} is a functional of $\tilde{V}$ and $V$ alone, one trained model serves every transfer function, colormap, and lighting configuration without re-encoding. Because $T$ is evaluated at sample positions rather than once per primitive, classification granularity follows the sampling rate rather than primitive extent, and because $\tilde{V}$ is continuous everywhere, derived quantities such as the shading gradient $\nabla \tilde{V}$ are available for shading at the sample position.

\begin{table}[t]
\raggedright
\caption{Implemented pipeline stages. Each is separately timed and memory-profiled; the analytic stages S1a--S1e together constitute allocation and complete in seconds even at billion-voxel scale.}
\label{tab:stages}
\begin{tabular}{ll}
\toprule
Stage & Operation \\
\midrule
P    & Load raw volume, normalize to $[0,1]$ \\
S1a  & Gradient magnitude, slab-wise \\
S1b  & Budget draw (detail + coverage), off-grid jitter \\
S1c  & Group scale estimates (subsampled median) \\
S1d  & Per-seed structure tensor, tangent extraction \\
S1e  & Assemble scales, quaternions, values, weights \\
S1x  & Export analytic model (pre-refinement baseline) \\
S2   & Fixed-budget refinement (1{,}500 iterations) \\
S3   & Export full precision ($+$ optional 8-bit) \\
EV   & Sampled PSNR evaluation (500K points) \\
\bottomrule
\end{tabular}
\end{table}

\subsection{Structure-Aware Fixed-Budget Allocation}
\label{sec:allocation}

The complete budget of $N$ primitives is allocated analytically before refinement. A 55\% detail fraction is sampled in proportion to gradient magnitude over the occupied domain ($V>0.02$), emphasizing boundaries and high-variation features; the remaining 45\% coverage fraction is sampled uniformly over that domain. Both draws enforce distinct positions, ensuring the realized budget matches the specification by construction; if a sparse field lacks sufficient seed positions, the encoder details the shortfall (Section~\ref{sec:seedlimit}). The importance sampling criterion is modular, permitting alternative physical quantities to substitute for gradient magnitude. Rather than placing primitives directly at voxel centers, each chosen coordinate $c_i$ is independently jittered within a half-voxel neighborhood,
\begin{equation}
\mu_i = c_i + u_i, \qquad u_i \sim \mathcal{U}\!\left[-\tfrac{1}{2}, \tfrac{1}{2}\right]^3 ,
\end{equation}


so decoupling primitive centers from grid alignment allow the representation to express off-grid structure. Each seed derives orientation and shape from a local $7^3$ structure tensor, aligning its principal axis with the local tangent, given by the eigenvector associated with the smallest eigenvalue. Axis scales are initialized from within-group seed spacing---tighter for detail seeds and broader for coverage seeds---then elongated along the tangent in proportion to gradient magnitude and clamped by a resolution-aware scale cap. Initial scalar attributes are sampled from the volume at seed positions, with uniform low initial weights. The allocated set is therefore a renderable, transfer-function-independent pre-refinement baseline. This one-pass initialization requires no subsequent splitting or pruning, fixing the model footprint before prior to refinement.


\subsection{Training Objective}
\label{sec:objective}

Training jointly refines all primitive parameters $(\mu_i, s_i, q_i, a_i, w_i)$ jointly with Adam~\cite{kingma2015adam}, supervised directly by the scalar field. Each iteration draws a stratified set of sample points: 35\% from high-value voxels ($V > 0.4$), 40\% from occupied mid-range voxels, and 25\% uniformly over the domain; all samples are jittered off-grid. Let $\tilde{V}(p)$ denote the reconstructed field from Eq.~\ref{eq:field} and $\rho(p) = \sum_i w_i G_i(p)$ its density. The objective is
\begin{equation}
\mathcal{L}
=
\mathcal{L}_{\mathrm{val}}
+
\lambda_{\rho}\,\mathcal{L}_{\rho}
+
\lambda_{\mathrm{sm}}\,\mathcal{L}_{\mathrm{sm}}
+
\lambda_{\mathrm{anc}}\,\mathcal{L}_{\mathrm{anc}}
+
\lambda_{\mathrm{a}}\,\mathcal{L}_{\mathrm{aniso}}
+
\lambda_{\mathrm{s}}\,\mathcal{L}_{\mathrm{cap}} .
\end{equation}

The value term minimizes reconstruction error at occupied samples:
\begin{equation}
\mathcal{L}_{\mathrm{val}} = \big\| \mathbf{1}[V(p) > 0.02] \odot \big(\tilde{V}(p) - V(p)\big) \big\|_2^2 .
\end{equation}
The density term drives $\rho(p)$ toward a uniform unit target over occupied space---applied bidirectionally to resolve both spatial gaps and overlap accumulations---leaving opacity response to the transfer function. The smoothness term minimizes density differences between nearby jittered pairs, ensuring spatial continuity instead of high-frequency speckles, its weight annealing from strong to weak so early iterations establish coverage and late ones recover detail. The anchor term ties each scalar to the volume value at the primitive's position; the anisotropy and scale-cap terms bound elongation at $3{:}1$ and evaluates scales outside the initialization-derived range, permitting sub-voxel scales and guarding against collapse. Training runs for a fixed 1{,}500 iterations across all budget ladders, with Section~\ref{sec:crossgpu} evaluating a reduced-iteration, time-budgeted variant. As membership in $\mathcal{G}$ remains invariant, training-based refinement updates positions, scales, orientations, scalar attributes, and weights while preserving primitive count and exported model size.

\subsection{Truncation-Aware Training at Gigavoxel Scale}

Naive evaluation of Eq.~\ref{eq:field} considers every primitive--sample pair. At $N=650{,}000$ primitives and 3{,}072 samples per iteration, this yields $2.0\times10^9$ candidate pairs, of which only 0.034\% contribute under the $5\sigma$ kernel truncation. We therefore bin primitives by their truncation radius and bucket sample points into grids matched to each bin's maximum radius. Evaluation is regularized to primitives in a $3^3$-cell neighborhood, followed by a bounding-sphere test which removes residual noncontributing pairs. As omitted Gaussian terms terms are identically zero, the procedure preserves the truncated field evaluation and its gradients up to float32 round-off. In the 650{,}000-primitive $1024^3$ benchmark, it reduced encoding time from 218.8 to 4.25 minutes ($51\times$) and peak refinement memory from 20.5 to 8.4\,GiB. Within the training loop, Gaussian contributions are accumulated in chunks with gradient checkpointing, bounding the active autograd graph by chunk size in place of the total pair count; while chunking divides computation, checkpointing prevents graph accumulation, enabling million-primitive budgets under a fixed memory ceiling. Gigavoxel volumes also demand memory-bounded analysis---slab-wise gradients, block-wise structure tensors, and a two-level budget draw that bypasses the allocation of full-volume index tensors---extending deployment beyond dedicated HPC clusters to a single 24\,GB commodity GPU for volumes exceeding a billion voxels and budgets up to at least $1.4$ million primitives.

\subsection{Implementation}
\label{sec:implementation}

The encoder is implemented in PyTorch with CUDA as an instrumented sequence of stages (Table~\ref{tab:stages})—load, analytic allocation, pre-refinement export, refinement, final export, and evaluation—each independently timed and memory-profiled. Two execution paths share this architecture: a direct path holding the volume resident, and a memory-bounded path for gigavoxel data. Both paths share the field evaluator, objective, and export pipeline, ensuring path-invariant execution; the truncation-aware evaluator carries a self-test validating values and gradients against dense evaluation. Runs are specified by the dataset configuration (detailed in Section~\ref{sec:setup}) and operational settings for memory and primitive budget, without dataset-specific tuning heuristics.

\subsection{Checkpointing, Export, and Rendering}

Periodic checkpoints hold the full primitive state and accumulated timing, allowing interrupted runs to resume with its reported cost without repeating allocation. The final representation stores twelve float32 values per primitive (position, log-scale, quaternion, scalar, weight) for 48 bytes per primitive, plus a small fixed header carries volume metadata and default lookup tables, facilitating interactive viewing while remaining distinct from the underlying representation. Rendering reconstructs the field of Eq.~\ref{eq:field} and applies the active transfer function through two execution paths that evaluate the identical encoded set: Gaussian splatting~\cite{zwicker2001ewa} and mixture-field ray marching. This separation provides a diagnostic framework which serves to distinguish data-level representation issues from renderer-specific artifacts through the evaluation stage.

\subsection{Compression and Storage Cost}
\label{sec:compression}

Our method replaces the dense grid with a compact set of Gaussian primitives rather than entropy-coding voxel residuals. A source volume with $M$ voxels stored at $b$ bytes per voxel occupies $S_{\mathrm{src}}=Mb$ bytes. The full-precision encoding stores $N$ primitives, each with twelve float32 values---three position values, three log-scales, four quaternion values, one scalar attribute, and one weight---leading to a total encoded size of
\begin{equation}
S_{\mathrm{enc}}(N) \;=\; 48\,N \;+\; H
\label{eq:size}
\end{equation}
bytes, where $H$ is a fixed header containing volume metadata and default transfer-function lookup tables. With $H$ independent of $N$, the compression ratio evaluates to
\begin{equation}
\mathrm{CR} \;=\; \frac{S_{\mathrm{src}}}{S_{\mathrm{enc}}(N)} \;\approx\; \frac{M b}{48\,N},
\label{eq:cr}
\end{equation}
depending exclusively on the source volume and the chosen budget. In contrast to optimization-driven encoders that yield variable output sizes, $\mathrm{CR}$ is determined a priori once $N$ is selected; targeting a specific compression ratio reduces directly to solving Eq.~\ref{eq:cr} for $N$. Morerover,  the storage footprint remains content-agnostic, ensuring a $200{,}000$-primitive encoding occupies an identical byte budget regardless of field complexity. The optional 8-bit format requires approximately 12 bytes per primitive, increasing the compression ratio by approximately $4\times$; both formats remain independent of transfer-function state. Reported ratios use the source byte count at each dataset's evaluated resolution and native storage format (Table~\ref{tab:datasets}).


\subsection{Relation to Prior Approaches}

Our method similarly uses field-space supervision by quering scalar values at sampled 3D positions, while the departure from prior scalar-aware Gaussian encodings lies in capacity allocation~\cite{dyken2025veg,dyken2026wveg}. W-VEG grows and prunes a partial set toward a ratio-derived target count; here the complete set is allocated analytically before optimization and preserved at exact count throughout training. The full budget is placed in a single analytic pass, committing capacity from field structure alone prior to any error signal. The count---and the encoded size of Eq.~\ref{eq:size}---is set from allocation through export. Initialization is anisotropic and structure-aware, so orientation and extent carry field information from the first iteration. Training consists purely of parameter refinement, so every iteration spends its cost on fitting the field rather than on steering a growth process. And the analytic stage alone yields a valid, renderable encoding, so refinement may stop at any point and still leave a complete model. Training focuses solely on optimizing field parameters without primitive management overhead; as the analytical stage alone produces a renderable encoding, refinement may be interrupted at any point while preserving the encoded model.

\subsection{Scope and Design Trade-offs}
\label{sec:tradeoffs}

When a user specifies interest in emphasizing a particular visual density feature, an optional render-time density mapping may be tuned against reference renders from user-selected viewpoints without modifying encoded primitives. Field fidelity remains preserved while updating only render-time mapping, requiring a small fraction of the initial encoding cost. Experimental boundaries and evaluation parameters specific to the core method are addressed separately in Section~\ref{sec:limitations}.

%% file: sections/04_experimental_setup.tex
\begin{table}[h]
\raggedright
\caption{Datasets. Source size is the distributed form at the evaluated resolution (stride-2 where marked), which all compression ratios and PSNR values are measured against. Occupancy is the fraction of voxels with normalized value above 0.02.}
\label{tab:datasets}
\footnotesize
\begin{tabular}{lccc}
\toprule
Dataset & Native resolution & Source (eval.) & Occ. \\
\midrule
Vortex & $128^3$ & 8.4\,MB f32 & 99.6\% \\
Bubble Plume & $256{\times}256{\times}640^{\,\dagger}$ & 21.0\,MB f32 & 2.4\% \\
Miranda & $1024^3$ & 4{,}295\,MB f32 & 95.1\% \\
Chameleon & $1024{\times}1024{\times}1080$ & 4{,}530\,MB f32 & 8.7\% \\
Richtmyer--Meshkov & $2048{\times}2048{\times}1920^{\,\dagger}$ & 1{,}007\,MB u8 & 51.6\% \\
\bottomrule
\multicolumn{4}{l}{\footnotesize $^{\dagger}$Evaluated at stride 2 per axis.}
\end{tabular}
\end{table}

\section{Experimental Setup}
\label{sec:setup}

\subsection{Datasets}

Experiments use five scientific scalar datasets whose evaluated grids span 2.1 million to 1.1 billion voxels (Table~\ref{tab:datasets}; native resolutions reach $2048{\times}2048{\times}1920$): Vortex ($128^3$), a structured flow field; Bubble Plume ($256{\times}256{\times}640$, stride-2), a sparse multiphase flow with 2.4\% occupancy; Miranda ($1024^3$)~\cite{klacansky2017open}, a hydrodynamic instability field with spatially distributed complexity; Chameleon ($1024{\times}1024{\times}1080$, float32), a CT scan with 8.7\% occupancy; and Richtmyer--Meshkov ($2048{\times}2048{\times}1920$), a turbulent-mixing simulation~\cite{klacansky2017open} --- evaluated at stride-2 as the full-resolution field (32.2\,GB as float32) exceeds the 24\,GB evaluation GPU---its complexity concentrates on a thin mixing interface, providing the interface-dominated case studied in Section~\ref{sec:structure}.

\begin{figure}[h]
\raggedright 
\includegraphics[width=\linewidth]{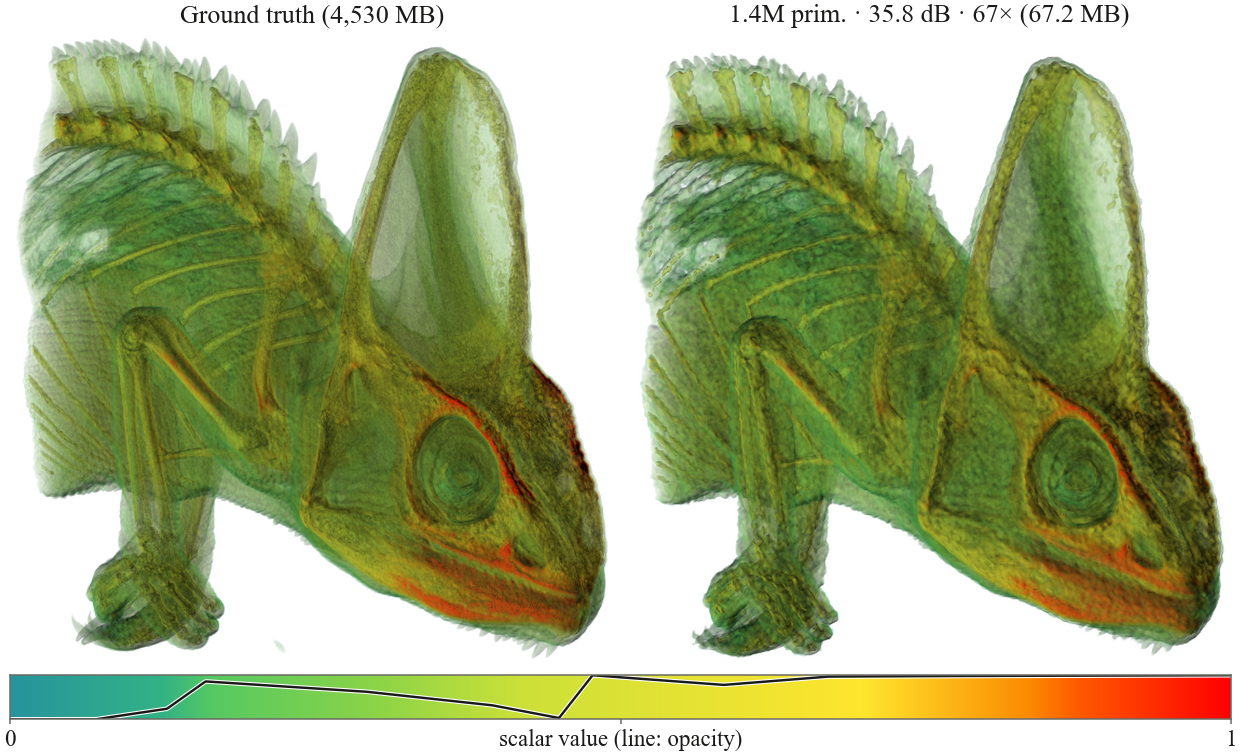}
\caption{Chameleon (8.7\% occupancy); identical transfer function, camera,
and lighting in both panels. Sparse concentrated fields convert budget most
efficiently: this ladder gains 16.5 dB and is still improving at the largest
budget.}
\label{fig:chameleon_full}
\end{figure}

\begin{figure}[t]
\raggedright 
\includegraphics[width=1.05\linewidth]{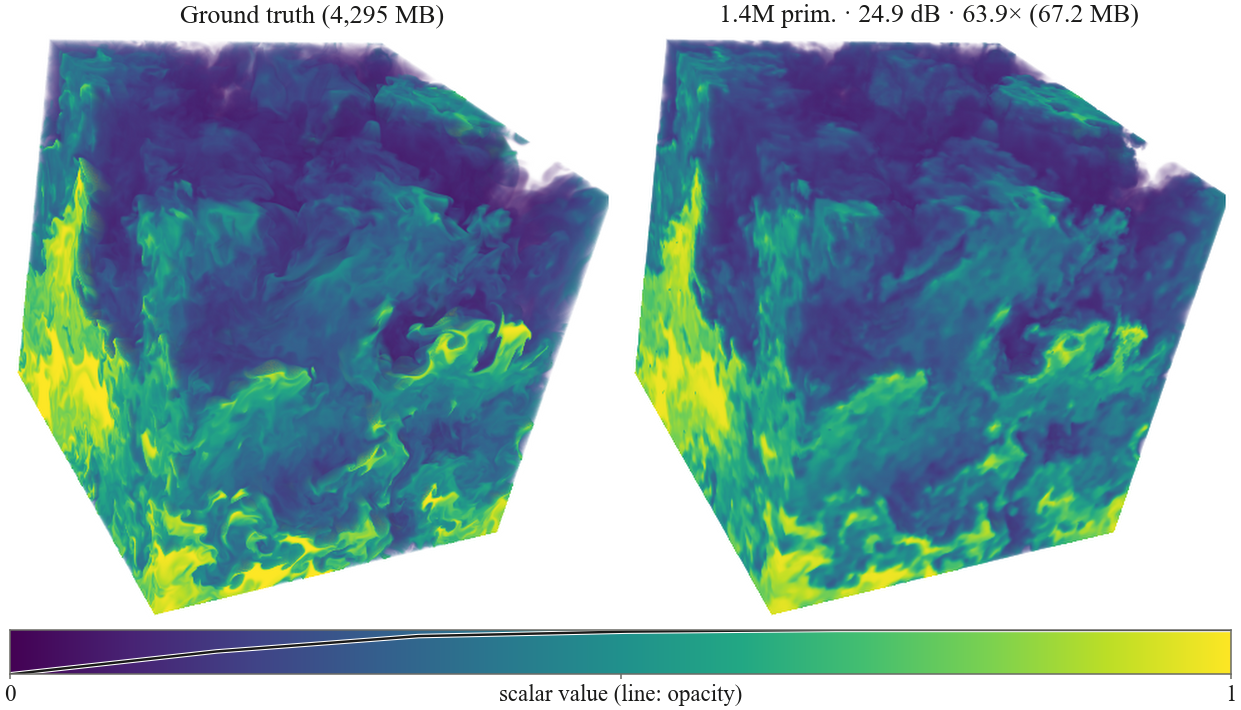}
\caption{Miranda, whose complexity is distributed nearly uniformly; identical transfer function and camera in both panels. Where complexity is distributed, one-shot allocation converts budget into quality steadily.}
\label{fig:miranda}
\end{figure}

\subsection{Training Configuration}

All budget ladders use the common configuration in Table~\ref{tab:training-config}. Budgets range from $10^3$ primitives on the smallest volumes to $1.4\times10^6$ primitives on gigavoxel volumes. The 48-byte full-precision format ceases to provide compression when its size exceeds the source volume; larger Vortex budgets are nevertheless included to characterize the rate--fidelity trade-off beyond this threshold. By default, the budget ladders in Table~\ref{tab:ladders} and costs in Table~\ref{tab:cost} were measured using one NVIDIA RTX~4090 GPU (24\,GB) under PyTorch/CUDA. Section~\ref{sec:crossgpu} additionally reports results on an A100 as a datacenter-class reference.

\subsection{Evaluation Metrics}

We evaluate reconstruction quality via volumetric PSNR, positioned as the standard metric for learned volume encoders~\cite{dyken2026wveg,lu2021compressive}, defined as $10\log_{10}(1/\mathrm{MSE})$ over evaluation positions where the $[0,1]$ normalization of Section~\ref{sec:method} fixes $V_{\max}=1$. PSNR is computed at 500{,}000 uniformly sampled voxel centers using a fixed evaluation seed and the pipeline's native evaluator, ensuring exact metric-encoder semantic alignment; sampled estimates match full-grid evaluations within 0.01\,dB. Because evaluation operates directly on scalar field values, the metric is strictly independent of transfer functions or renderers. PSNR weights every position equally regardless of visibility or perceptual salience---providing a rigorous, visualization-independent reference for encoded field accuracy rather than a prediction of rendered fidelity~\cite{wang2009mse,wang2004ssim}. On majority-occupied fields, restricting PSNR to non-empty voxels equals or exceeds full-field values by 0.04--1.5\,dB, confirming that reported figures reflect true material data rather than background padding; full-field metrics are reported throughout. Total training time sums all instrumented pipeline stages, and peak GPU memory is profiled per stage. Two operational parameters define execution: first, stochastic allocation and refinement sampling induce a minor 0.2--0.5\,dB spread across repeated runs (reported where applicable); second, all timing figures reflect the $51\times$ accelerated, truncation-aware implementation, with quality verified to be hardware-independent (Section~\ref{sec:crossgpu}).

\begin{table}[h]
\raggedright 
\caption{Core training configuration, identical across datasets.}
\label{tab:training-config}
\footnotesize
\begin{tabular}{lc}
\toprule
Setting & Value \\
\midrule
Training iterations & 1{,}500 \\
Sample points / iteration & 3{,}072 (stratified 35/40/25) \\
Optimizer & Adam, lr $8\times10^{-3}$ \\
Allocation split & 55\% gradient-weighted / 45\% uniform \\
Kernel truncation & $5\sigma$ \\
Anisotropy bound & $3{:}1$ \\
Storage & 48 B/primitive (+ optional 8-bit) \\
Hardware & RTX 4090 desktop GPU, 24\,GB \\
\bottomrule
\end{tabular}
\end{table}

%% file: sections/05_results.tex
\begin{figure}[b]
\raggedright 
\includegraphics[width=1\linewidth]{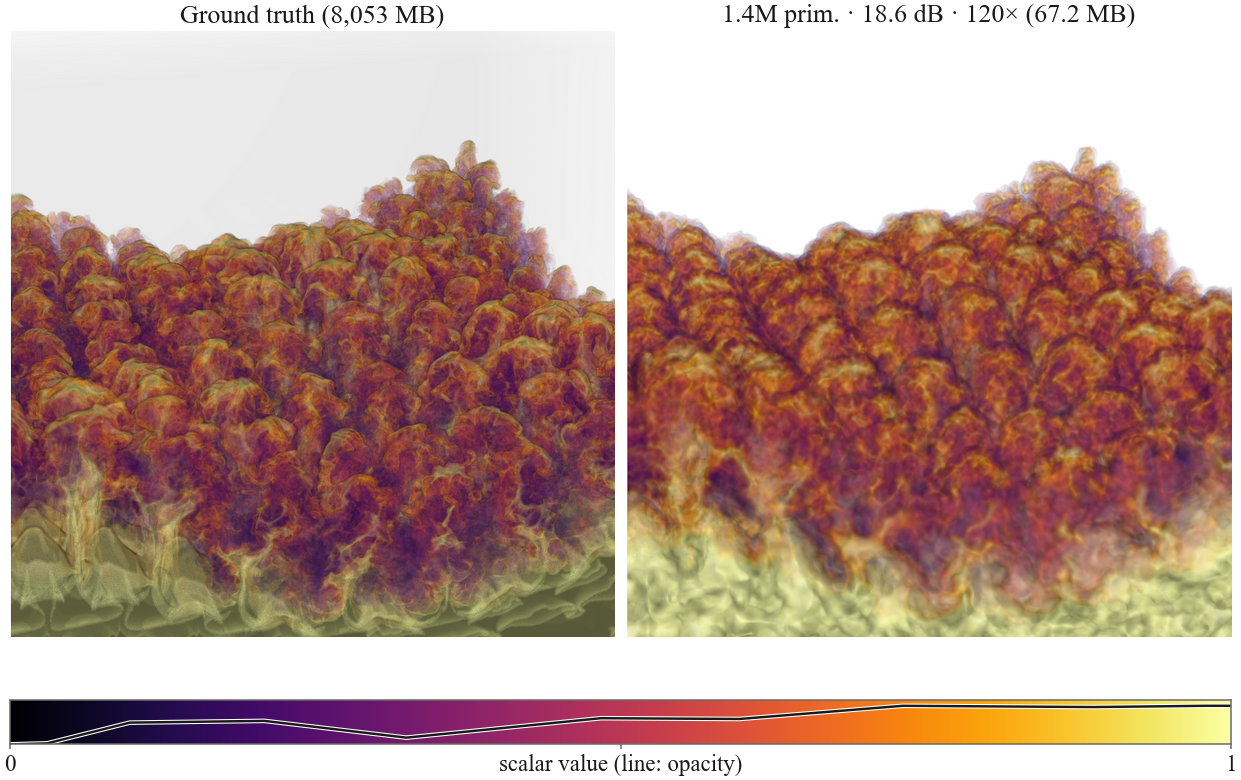}
\caption{Richtmyer--Meshkov; identical transfer function and camera in both panels. Its gradient mass concentrates on the thin mixing interface (77.7\% in the top decile of occupied voxels), the region Section~\ref{sec:structure} identifies in advance.}
\label{fig:richtmyer}
\end{figure}

\section{Results}

The evaluation investigates reconstruction quality across budgets, the budget ranges a field admits, encoding cost (including a reduced-iteration time budget), and pre-encoding structure indicators of fields for which one-shot allocation yields limited gains from additional capacity.

\begin{table}[b]
\raggedright 
\caption{Reconstruction PSNR (dB) across primitive budgets; dashes mark budgets outside a dataset's ladder. The bottom row gives the encoded size at each budget: it follows from Eq.~\ref{eq:size} alone and is therefore identical for every dataset, known before training begins. All encodings use the configuration of Table~\ref{tab:training-config}; compression follows Eq.~\ref{eq:cr} from the budget and the source bytes of Table~\ref{tab:datasets}.}
\label{tab:ladders}
\setlength{\tabcolsep}{2.5pt}
\footnotesize
\begin{tabular}{lccccccccc}
\toprule
 & \multicolumn{9}{c}{PSNR (dB) at primitive budget} \\
\cmidrule(lr){2-10}
Dataset & 2K & 10K & 20K & 32K & 90K & 200K & 650K & 1M & 1.4M \\
\midrule
Vortex & 21.0 & 24.7 & 26.8 & 28.4 & 32.9 & 36.9$^{\S}$ & 41.9$^{\S}$ & --- & 44.9$^{\S}$ \\
Bubble & 27.2 & 32.1 & 33.5 & 34.6 & 37.3 & 38.7 & $\ast$ & $\ast$ & $\ast$ \\
Miranda & 18.0 & 19.7 & 19.9 & 20.1 & 20.9 & 21.9 & 23.6 & 24.3 & 24.9 \\
Chameleon & 19.2 & 21.4 & 22.8 & 24.1 & 26.8 & 29.2 & 33.1 & 34.6 & 35.7 \\
Richtmyer & 15.0 & 15.5 & 15.6 & 15.7 & 16.5 & 17.1 & 18.1 & 18.4 & 18.6 \\
\midrule
Size (MB) & 0.10 & 0.49 & 0.97 & 1.54 & 4.33 & 9.61 & 31.2 & 48.0 & 67.2 \\
\bottomrule
\end{tabular}

\vspace{3pt}
{\footnotesize \quad
$^{\dagger}$Vortex additionally reaches 19.8\,dB at $153\times$ from 1{,}000 primitives.\quad
$^{\S}$Encoded model exceeds its own source volume ($\mathrm{CR}<1$); reported for completeness, outside the compression-useful range.\quad
$\ast$Budget exceeds the field's distinct structured positions and is not attainable; see Section~\ref{sec:seedlimit}.}
\end{table}

\subsection{Reconstruction Quality Across Budgets}
\label{sec:quality}
\begin{figure}[h]
\raggedright 
\includegraphics[width=\linewidth]{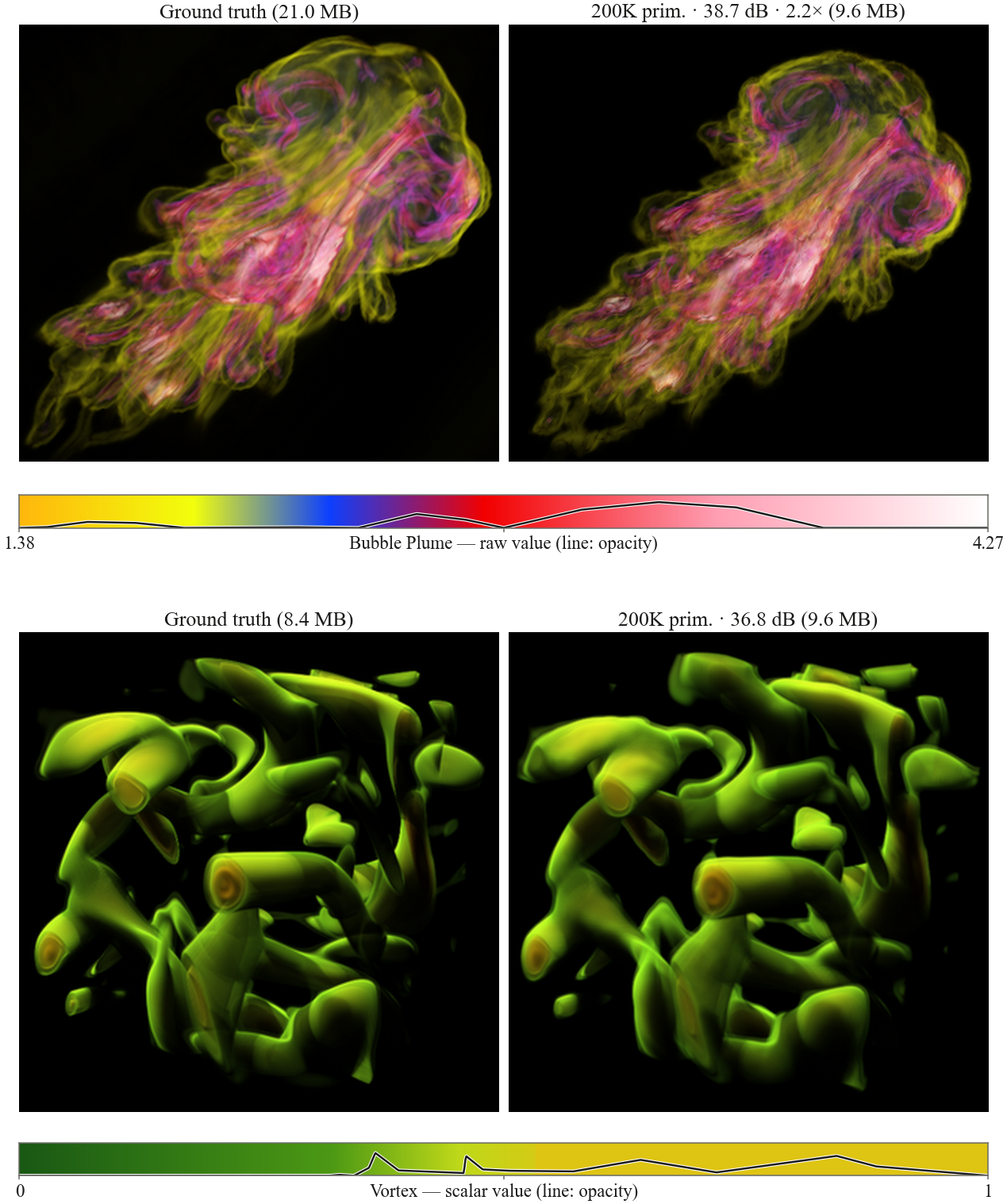}
\caption{The two smaller datasets at 200{,}000 primitives each; identical transfer function and camera per pair, transfer function shown beneath. Vortex is at near size parity (0.9$\times$, Section~\ref{sec:seedlimit}) at 36.9\,dB. The pair spans the occupancy extremes (2.4\% and 99.6\%), which set their budget ceilings (Section~\ref{sec:seedlimit}).}
\label{fig:small}
\end{figure}

\begin{figure}[h]
\raggedright 
\includegraphics[width=\linewidth]{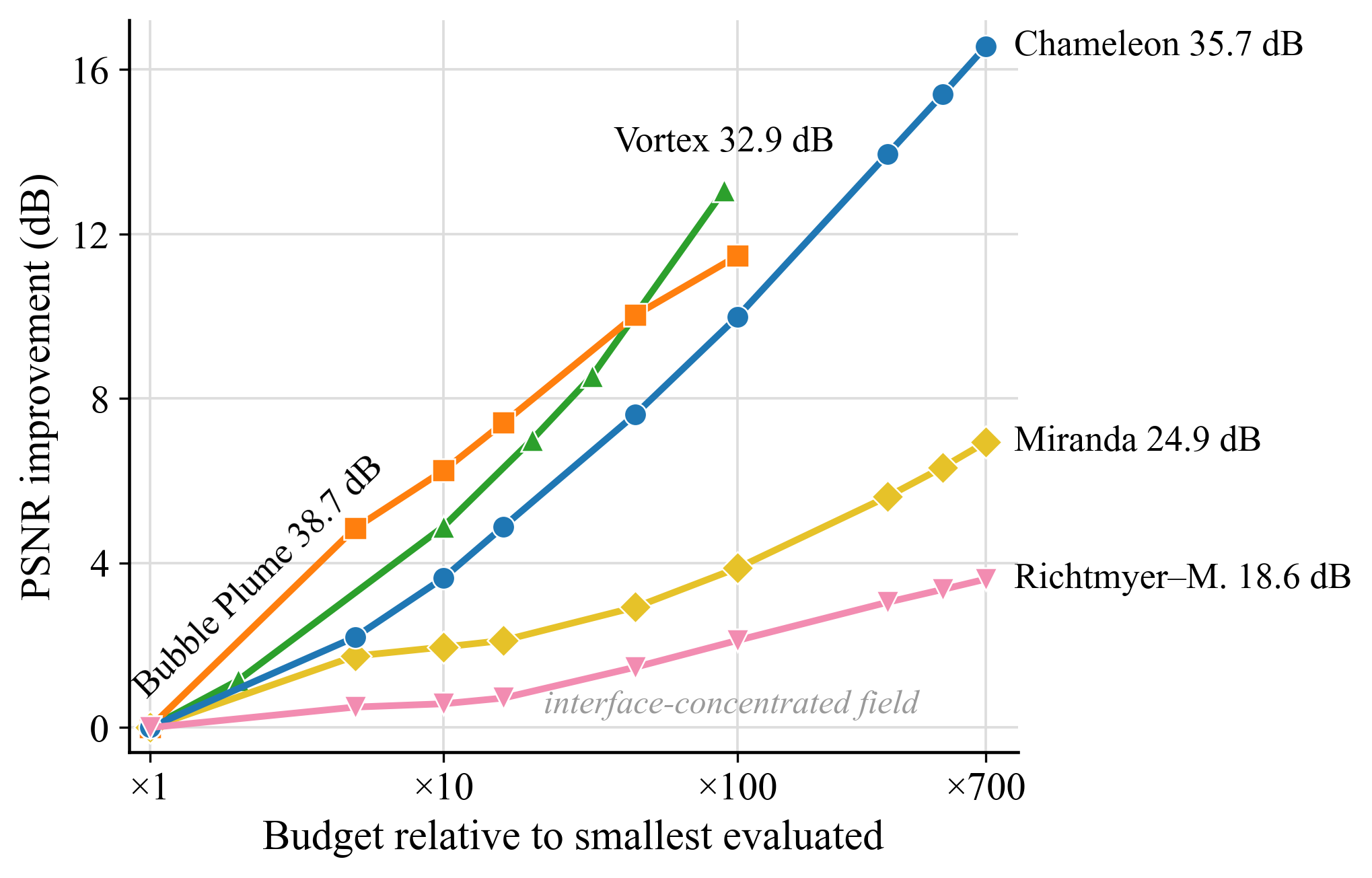}
\caption{Quality improvement against relative budget, each ladder plotted against its own smallest budget ($\times$1) over its compression-useful range; end labels give final absolute PSNR. Four datasets convert budget into 6.9--16.5\,dB; Richtmyer--Meshkov stands apart at $+3.6$\,dB, the case Table~\ref{tab:structure} identifies in advance. Absolute levels differ because fields differ in complexity; the improvement curves are the comparable signal.}
\label{fig:ladders}
\end{figure}

Table~\ref{tab:ladders} reports reconstruction PSNR across budget ladders for all five datasets. The three gigavoxel fields---Miranda, Chameleon, and Richtmyer--Meshkov (Figs.~\ref{fig:chameleon_full}--\ref{fig:richtmyer})---represent our primary target workloads for in situ reduction, while Vortex and Bubble Plume span spatial occupancy extremes. Three patterns emerge: sparse, structure-concentrated fields convert budget most efficiently --- Bubble Plume (Fig.~\ref{fig:small}) reaches 27.2\,dB at $204\times$ compression from 2{,}000 primitives and 38.7\,dB at 200{,}000 primitives, and Chameleon (Figs.~\ref{fig:chameleon_full} \&~\ref{fig:chameleon}) gains 16.5\,dB across its ladder, continuing to improve at 1.4~million primitives. Uniformly complex Miranda (Fig.~\ref{fig:miranda}) gains 6.9\,dB over a 700-fold budget range with accelerating per-doubling returns. Turbulent Richtmyer--Meshkov (Fig.~\ref{fig:richtmyer}) yields a modest 3.6\,dB improvement while achieving $9{,}773\times$ compression at its smallest budget---a behavior predicted directly from field analysis in Section~\ref{sec:structure}. Figures~\ref{fig:chameleon_full}--\ref{fig:richtmyer} illustrate each gigavoxel dataset at its highest evaluated budget against dense ground truth, whereas Fig.~\ref{fig:small} depicts the two smaller datasets; intermediate allocations appear in Table~\ref{tab:ladders} and Fig.~\ref{fig:ladders}. Each pair shares an identical transfer function, camera angle, and lighting configuration. The encoded panel additionally applies the render-time density mapping of Section~\ref{sec:tradeoffs}---a render-side opacity control, distinct from the transfer function, that leaves underlying primitives untouched and all field-space metrics unchanged. All visualization settings are serialized as viewer configuration files to guarantee exact reproducibility, with PSNR (Section~\ref{sec:setup}) serving as the quantitative metric throughout. Compression ratios follow directly from Eq.~\ref{eq:cr}: at 1.4~million primitives, gigavoxel encodings reach $64\times$ (Miranda), $67\times$ (Chameleon), and $15\times$ (Richtmyer--Meshkov, evaluated byte basis); at minimal allocation (2{,}000 primitives occupying 103\,kB), representations are significantly compact, reconstructing Chameleon at $43{,}979\times$, Miranda at $41{,}699\times$, and Richtmyer--Meshkov at $9{,}773\times$.

\subsection{Budget Limits Imposed by Field Structure}
\label{sec:seedlimit}

Two constraints, both known before optimization, determine the feasible and compression-beneficial budget range for each field. The first is storage: At 48 bytes per primitive, an encoding ceases to provide compression when its budget exceeds $S_{\mathrm{src}}/48$. This threshold is approximately 175{,}000 primitives for Vortex and 437{,}000 for Bubble Plume. Larger encodings remain valid and may continue to improve reconstruction quality: Vortex reaches 44.9\,dB at $1.4\times10^6$ primitives, the highest fidelity measured in our evaluation. Table~\ref{tab:ladders} reports these results separately because they exceed the corresponding source-volume sizes. The second constraint is structural and primarily affects sparse fields. Because detail seeds explicitly occupy distinct gradient-bearing voxels, Bubble Plume's 125{,}787 eligible voxels support approximately $2.3\times10^5$ primitives under the default 55\% detail allocation. Requests are validated upfront, ensuring the system returns a complete model matching the requested budget or provides an immediate status signal. By contrast, Vortex's 99.6\% occupancy supports the full $1.4\times10^6$-primitive budget, showing that the limit depends on the distribution of field structure rather than source size. Table~\ref{tab:ladders} confirms that primitive budgets set output footprints deterministically prior to optimization (Eq.~\ref{eq:size}), with 8-bit exports requiring approximately one quarter of the full-precision storage.

\subsection{Training Cost and Memory}
\label{sec:cost}

\begin{table}[t]
\raggedright 
\caption{Training cost on one RTX 4090 desktop GPU (representative rows; time is the instrumented pipeline total, memory the peak of the refinement stage).}
\label{tab:cost}
\footnotesize
\begin{tabular}{lccc}
\toprule
Dataset / budget & Time & Train mem. & Model \\
\midrule
Vortex 10K & 2.9 min & 1.7\,GiB & 0.49\,MB \\
Bubble 200K & 2.6 min & 4.0\,GiB & 9.6\,MB \\
Miranda 650K & 2.6 min & 8.4\,GiB & 31.2\,MB \\
Chameleon 1M & 2.3 min & 8.3\,GiB & 48.0\,MB \\
Miranda 1.4M & 3.8 min & 9.5\,GiB & 67.2\,MB \\
Richtmyer 1.4M & 4.0 min & 8.1\,GiB & 67.2\,MB \\
\bottomrule
\end{tabular}
\end{table}

Table~\ref{tab:cost} summarizes training cost on the RTX~4090. Representative encodings complete in at most 4.0 minutes, with peak refinement memory ranging from 1.7 to 9.5\,GiB; one-time analysis and seeding stages complete in seconds, including for gigavoxel inputs. These properties---seconds-to-minutes execution, deterministic output size, and bounded memory---support the feasibility of in situ reduction, while validation within a coupled simulation remains future work. Truncation-aware evaluation accounts for most of this scalability: at 650{,}000 primitives, only 0.034\% of the $2.0\times10^9$ candidate Gaussian--point pairs per training iteration are nonzero under the $5\sigma$ cutoff. Relative to dense evaluation, leveraging this sparsity reduced Miranda encoding time from 218.8 to 4.25 minutes ($51\times$) and peak memory from 20.5 to 8.4\,GiB, while matching field values to float32 round-off and gradients exactly. Because dense evaluation allocates 20.5\,GiB at 650{,}000 primitives, scaling to million-primitive budgets exceeds the capacity of standard 24\,GB GPUs. Implementation consistency and stochastic variance were assessed independently. Across software revisions, a 1-million-primitive Miranda encoding yielded PSNR agreement within 0.01\,dB, whereas unseeded reruns at 650{,}000 primitives produced a minor 0.22\,dB spread---consistent with the expected $\pm 0.2\text{--}0.5$\,dB run-to-run band.

\subsection{Cross-GPU Cost and a Time-Budget Mode}
\label{sec:crossgpu}

\begin{figure}[b]
\raggedright 
\includegraphics[width=\linewidth]{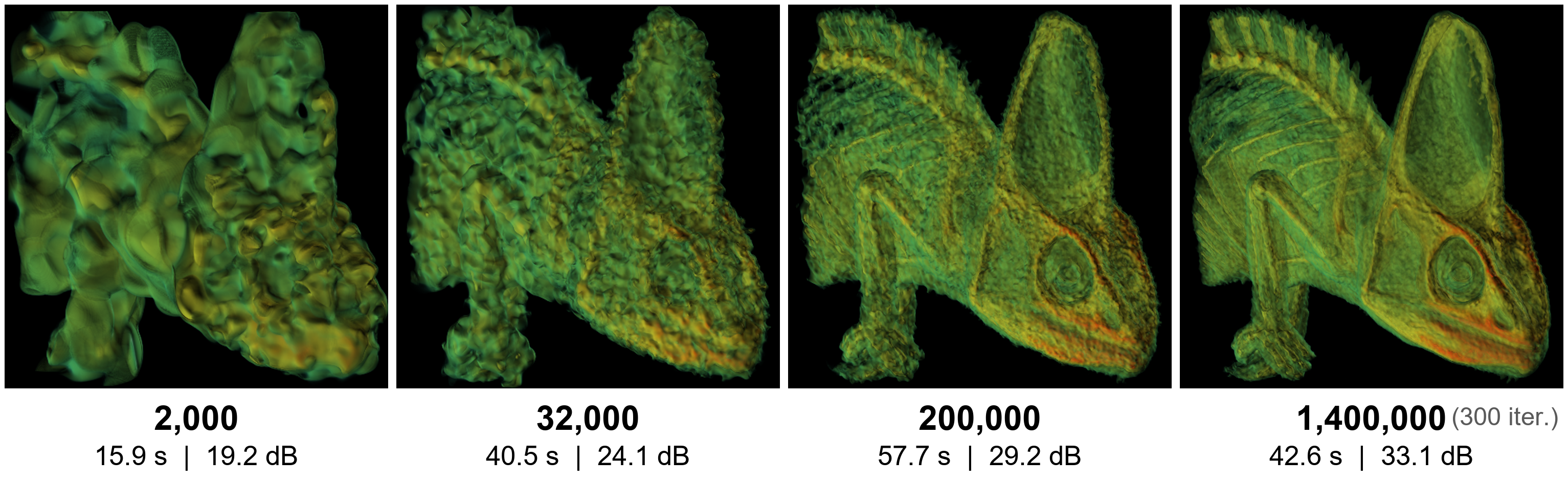}
\caption{Chameleon (8.7\% occupancy) across budgets, one transfer function and density setting throughout; per panel: primitives, RTX~4090 refinement seconds, and PSNR. The last panel is the reduced-iteration time-budget mode. Quality tracks the budget visibly, and the ladder gains 16.5\,dB overall.}
\label{fig:chameleon}
\end{figure}

Reconstruction results were consistent across the tested hardware and implementation paths. Across 21 configurations spanning three gigavoxel datasets and nine budgets, encodings produced on a second GPU with an independent reference implementation agreed with Table~\ref{tab:ladders} to within 0.17\,dB (mean 0.08\,dB). Execution runtime, however, varied by architecture: at 1.4 million primitives and 1{,}500 iterations, Miranda refined in 201.5\,s on the RTX~4090 versus 280.5\,s on an A100, while Richtmyer--Meshkov required 186.9\,s and 247.7\,s, respectively---rendering the RTX~4090 $1.33$--$1.39\times$ faster on this workload. Because encoding requires neither cameras, transfer functions, nor rendered images (Section~\ref{sec:method}), these computational costs and the resulting reconstruction fidelity remain strictly independent of viewpoint and transfer-function choices; a single encoding run serves every subsequent visualization state. Furthermore, because budget---and thus encoded size---is fixed by construction, iteration count acts as a direct regulator of optimization latency. At 300 iterations, the same 1.4-million-primitive models refine in 52.8\,s (Miranda), 42.6\,s (Chameleon, Fig.~\ref{fig:chameleon}), and 44.2\,s (Richtmyer--Meshkov) on the RTX~4090, retaining strong quality within 1.3--2.7\,dB of full convergence using a closed-form-gradient formulation verified against the reference implementation before each run. On identical hardware, 1.4 million primitives at 300 iterations matches the reconstruction fidelity of 650{,}000 primitives at 1{,}500 iterations in less than half the time. Under a fixed time constraint, opting for a larger primitive capacity with fewer iterations provides the superior operating regime: a billion-voxel field reduced to a fixed 67.2\,MB with refinement under a minute, running on hardware a visualization group already owns.

\subsection{Rendering the Encoded Model}

The encoded models render interactively in the viewer through both rendering paths of Section~\ref{sec:implementation}, alongside VTK-based ray casting of the dense grid~\cite{schroeder2006vtk}. Frame rate is governed by the primitive count rather than by which training engine produced the model: the exported format and count are identical, so a model refined with the closed-form-gradient engine renders exactly as its reference-engine counterpart. Rendering cost therefore scales with the chosen budget---the same knob that sets encoded size---rather than with the source resolution. A controlled rendering-performance comparison under matched settings is left to future work.

\begin{table}[t]
\raggedright 
\caption{Pre-encoding structure statistics and measured ladder improvement ($\Delta$PSNR: dB gain across each dataset's compression-useful budget range; larger is better). Gradient concentration tracks $\Delta$PSNR, and Richtmyer--Meshkov---the one field with interface-concentrated gradient mass---is identified in advance.}
\label{tab:structure}
\footnotesize
\begin{tabular}{lcccc}
\toprule
Dataset & Occ. & Grad.\ top-10\% & $S(1)/S(16)$ & $\Delta$PSNR \\
\midrule
Vortex & 99.6\% & 25.0\% & 0.12 & $+13.1$ \\
Bubble Plume & 2.4\% & 28.4\% & 0.24 & $+11.5$ \\
Miranda & 95.1\% & 39.8\% & 0.10 & $+6.9$ \\
Chameleon & 8.7\% & 35.5\% & 0.10 & $+16.5$ \\
Richtmyer--M. & 51.6\% & \textbf{77.7\%} & 0.22 & $+3.6$ \\
\bottomrule
\end{tabular}
\end{table}

\subsection{Predicting Allocation Sufficiency from Field Structure}
\label{sec:structure}

Across 21 configurations spanning three gigavoxel datasets and nine budgets, results obtained on an RTX~5090 using an independent implementation differed from Table~\ref{tab:ladders} by at most 0.17\,dB (mean 0.08\,dB). Reconstruction quality was therefore effectively independent of the tested hardware and implementation path; runtime, by contrast, varied across GPUs. At 1.4~million primitives and 1{,}500 iterations, Miranda refined in 201.5\,s on the RTX~4090 and 280.5\,s on the A100, while Richtmyer--Meshkov required 186.9 and 247.7\,s, respectively. The tested RTX~4090 was therefore $1.33$--$1.39\times$ faster on these workloads than the A100, the hardware class on which prior scalar-aware Gaussian encodings report training cost~\cite{dyken2026wveg}. Reducing the iteration count provides a direct runtime--quality trade-off. At 300 iterations, the 1.4-million-primitive models refined in 52.8\,s for Miranda, 42.6\,s for Chameleon (Fig.~\ref{fig:chameleon}), and 44.2\,s for Richtmyer--Meshkov, with a 1.3--2.7\,dB reduction relative to 1{,}500 iterations. The closed-form gradient implementation matched the reference values and gradients to float32 round-off; its performance benefit varied across the tested GPUs. In the matched comparison, 1.4~million primitives at 300 iterations achieved the quality of 650{,}000 primitives at 1{,}500 iterations in less than half the time. This identifies a favorable time-constrained operating point: a gigavoxel field represented in 67.2\,MB with refinement completed in under one minute on the tested RTX~4090. Because the encoding is viewpoint- and transfer-function-independent (Section~\ref{sec:method}), this cost is incurred once per field. Coupled-simulation experiments remain necessary to validate its in situ performance.

%% file: sections/06_discussion.tex
\section{Discussion}

\subsection{Representation Behavior}

We demonstrate that volumetric fields can be approximated using explicit Gaussian primitives allocated in a single structure-aware pass under a fixed budget. Each primitive remains individually addressable, supporting primitive-level selection and adaptive processing, while primitive count and encoded size remain fixed specifications rather than outcomes of optimization. As each primitive retains a scalar attribute rather than baked RGB appearance, the encoding remains a scalar-field representation: transfer functions and colormaps can be changed, and lighting recomputed, without re-encoding. A single trained model thus supports unconstrained post-hoc exploration, bringing the transfer-function flexibility once reserved to image-space proxies~\cite{tikhonova2010explorable} directly into a compact, fully 3D field representation.

\subsection{Limitations}
\label{sec:limitations}

The principal limitation arises from one-shot allocation: because capacity is committed before reconstruction error is available, the resulting primitive distribution may leave room for further spatial refinement when field complexity is concentrated within narrow interfacial regions. Section~\ref{sec:structure} identifies this condition before training, and the capacity experiments indicate that allocation---rather than total budget or initial primitive size---is the likely bottleneck. Preliminary experiments suggest that render-time density tuning can reduce visual discrepancies without modifying the encoded primitives, fixed memory budget, field-space reconstruction, or PSNR. While position refinement is driven continuously through the field-space objective, discrete primitive relocation---shifting low-contribution elements to high-error regions---offers a direct, budget-preserving solution. This extension serves as a complementary future direction that may improve field-space fidelity while preserving the exact primitive count, with the structure statistics serving as a selective threshold for execution. A controlled matched-platform comparison with prior encoders remains future work.

%% file: sections/07_conclusion.tex
\section{Conclusion}

We developed, implemented, and evaluated a fixed-budget Gaussian encoding framework for scientific scalar volume representation. The central methodological contribution is the analytic, structure-aware allocation of the complete anisotropic primitive ensemble prior to optimization, maintaining an exact primitive count throughout camera-free field-space refinement. For in situ data reduction, this transforms the encoded footprint from an empirical byproduct of optimization into a deterministic constraint aligned with fixed memory budgets. Across five datasets, fixed-count representations provide useful rate--fidelity trade-offs across sparse, dense, and interface-dominated fields, while truncation-aware evaluation ensures scalability at gigavoxel scale. Our method determines capacity from field structure a priori, establishing encoded size as a controllable specification rather than a byproduct of optimization-time growth. Capacity budget progression demonstrated that spatially localized fields benefited most from additional primitive capacity, while lightweight pre-encoding statistics successfully identified interface-dominated fields where one-shot allocation yields bounded capacity gains. Because primitives store underlying scalar attributes rather than pre-rendered appearance, a single representation supports post-hoc transfer-function, colormap, lighting, and viewpoint modifications without re-encoding. Collectively, deterministic memory footprinting, sub-10\,GB peak VRAM utilization, controllable optimization latency, and post-hoc visualization reuse establish the method as a practical candidate for in situ data reduction~\cite{antic2026}. More broadly, this explicit representation provides a scalable foundation for distributed high-resolution rendering and client-side scientific visualization~\cite{han2025distributed,niedermayr2024cinematic}. These findings motivate coupled-simulation validation and extension to time-varying fields, a primary objective for learned volume compression~\cite{han2026moeinr}.

\section{Acknowledgements}
This research used resources of the Argonne Leadership Computing Facility and is supported in part by the U.S. Department of Energy, Office of Science, Office of Advanced Scientific Computing Research (ASCR), for the project Argonne Base: Scientific Data Management and Visualization to Advance AI for Science, under Contract DE-AC02-06CH11357.